\documentclass{article}
\usepackage{iclr2026_conference,times}

\usepackage{amsmath,amsfonts,bm}

\def\eqref#1{equation~\ref{#1}}

\def\1{\bm{1}}

\DeclareMathAlphabet{\mathsfit}{\encodingdefault}{\sfdefault}{m}{sl}
\SetMathAlphabet{\mathsfit}{bold}{\encodingdefault}{\sfdefault}{bx}{n}

\usepackage{amsmath,amssymb}
\usepackage{booktabs}
\usepackage{graphicx}
\usepackage{microtype}
\usepackage{xcolor}
\usepackage{hyperref}
\usepackage{url}

\hypersetup{
  pdftitle={Evolutionary Algorithm-Guided LLMs for Physics-Informed Neural Network Design},
  pdfauthor={Xu Yang, Mingyang Yu, Jing Xu, Keqian Li}
}

\title{Evolutionary Algorithm-Guided LLMs for Physics-Informed Neural Network Design}

\author{
{\normalfont Xu Yang$^{1\dagger}$\quad Mingyang Yu$^{2\dagger}$\quad
Jing Xu$^{2}$\quad Keqian Li$^{1}$}\\
\normalfont\small
$^{1}$Shanghai Institute of Intelligent Education, East China Normal University, Shanghai 200062, China\\
$^{2}$College of Artificial Intelligence, Nankai University, Tianjin 300350, China\\
51290936034@stu.ecnu.edu.cn; 1120240312@mail.nankai.edu.cn\\
xujing@nankai.edu.cn; kqli@mail.ecnu.edu.cn\\
{\normalfont\normalsize \(\dagger\) These authors contributed equally to this work}
}

\iclrfinalcopy

\begin{document}
\maketitle
\lhead{}

\begin{abstract}
Physics-informed neural networks (PINNs) are unusually sensitive to interacting choices of architecture, activation, loss weighting, collocation, optimization, and constraint enforcement. Large language models (LLMs) can propose these choices, but independent recommendations do not accumulate experience from previously trained PINNs. We propose a closed-loop evolutionary algorithm that guides an LLM to generate complete, executable PINN configurations across generations, using measured training outcomes to determine subsequent search decisions. The algorithm maintains an evaluated population and lineage, applies parent-conditioned mutation and crossover, preserves elite and diverse solutions, rejects effective duplicates, and converts parent-relative successes and failures into the next-generation context supplied to the LLM. Every proposed configuration is executed directly under an exact optimizer-step budget. On a one-dimensional multiscale wave equation, two independent ten-generation runs trained 60 PINNs for 600,000 optimizer steps. In both runs, the best configuration appeared in the final generation, with best mean-squared error reduced by 2.97\% and 95.38\% relative to the initial population. The stronger run validated residual connections and increased depth on separate branches, combined them in a later generation, and then refined width and collocation density. It also revealed that low solution error can coexist with a high PDE residual. These results demonstrate the feasibility of evolutionary-algorithm-guided LLMs for PINN design on a controlled PDE while motivating broader, physics-aware evaluation.
\end{abstract}

\section{Introduction}

Physics-informed neural networks embed differential equations and physical constraints into neural-network training, offering a mesh-free route to forward and inverse scientific problems \citep{raissi2019pinn,karniadakis2021physics}. Their apparent simplicity hides a difficult algorithm-design problem. Network depth and width, activation functions, loss weights, collocation strategies, optimizers, learning-rate schedules, and hard or soft constraint mechanisms interact nonlinearly. Poor combinations can produce gradient imbalance, spectral bias, or optimization failure even when the governing equation is correctly specified \citep{wang2021gradient,wang2022ntk,krishnapriyan2021failure}. Consequently, selecting a PINN is not merely hyperparameter tuning over independent coordinates; it is the design of a coupled numerical learning algorithm.

LLMs provide a promising interface for this design problem because they can reason over structured descriptions, scientific priors, and heterogeneous algorithm components. Recent systems use LLMs for optimization \citep{yang2024opro}, program search \citep{romeraparedes2024funsearch}, and PDE surrogation \citep{wuwu2025pinnsagent}. Our question is not whether an LLM can name a good PINN configuration once. It is whether a purpose-built evolutionary architecture can use numerical outcomes to guide later LLM designs. In our initial development experiments, the first proposal could remain best while later suggestions repeated equivalent configurations or introduced unsupported changes. The missing ingredient was an explicit search state that makes numerical survival, ancestry, diversity, and failed hypotheses part of the next decision.

We therefore introduce an evolutionary architecture developed in this work specifically for PINN configuration search. It guides an LLM to design complete structured \emph{AlgorithmSpec} objects, each of which declaratively specifies the network, loss, sampler, optimizer, schedule, and physical constraints. Our architecture controls parent selection, lineage, elite survival, fingerprint deduplication, numerical evaluation, and cross-generation memory. The LLM serves as a conditional design operator that proposes mutations and crossovers from the state assembled by the architecture. Optimization is therefore performed by the proposed architecture rather than by a sequence of independent LLM recommendations.

Our contributions are:
\begin{itemize}
    \item We introduce a PINN-specific evolutionary architecture, proposed in this work, that guides an LLM to design complete structured PINN algorithms rather than acting as a one-shot tuner.
    \item We connect effective-specification fingerprinting, distinct-parent crossover, novelty-aware selection, elite-and-diversity updates, and success/failure reflection into a measured cross-generation feedback loop.
    \item We implement direct, auditable execution: the LLM does not emit executable Python, unsupported components fail without silent substitution, and every successful PINN is trained under an exact optimizer-evaluation budget.
    \item We provide a lineage-resolved Wave-C study in which both independent runs improve beyond their initial generation and the stronger run achieves a 95.38\% MSE reduction through a sequence of interpretable evolutionary changes.
    \item We identify objective mismatch as a concrete failure mode: the lowest-MSE model has a substantially larger PDE residual, motivating physics-guarded multi-objective evolution.
\end{itemize}

The present evidence is intentionally bounded. It demonstrates the search mechanism on one analytic PDE with two independent seeds; it does not establish state-of-the-art PINN performance across equations.

\section{Related Work}

\paragraph{Physics-informed neural networks.}
PINNs combine governing-law residuals with boundary, initial, and observation losses \citep{raissi2019pinn}. Subsequent work has shown that training quality depends strongly on gradient balance \citep{wang2021gradient}, neural-tangent-kernel dynamics \citep{wang2022ntk}, curricula and regularization \citep{krishnapriyan2021failure}, adaptive sampling \citep{wu2023sampling}, and exact constraint construction \citep{lagaris1998neural,sukumar2022boundary}. Libraries such as DeepXDE expose many of these choices \citep{lu2021deepxde}, while PINNacle demonstrates that no single configuration dominates all PDEs \citep{hao2024pinnacle}. Our work treats these coupled choices as an open structured algorithm-design space.

\paragraph{LLMs for optimization and algorithm discovery.}
Optimization by prompting uses objective values to iteratively improve natural-language candidates \citep{yang2024opro}. FunSearch combines LLM program generation with evolutionary evaluation \citep{romeraparedes2024funsearch}, while EoH represents heuristic ideas in language and evolves them through evolutionary operators \citep{liu2024eoh}. These studies establish useful precedents for coupling LLM generation with evaluated search history. Our method is not a direct implementation of EoH. It introduces a separate PINN-specific architecture built around structured full-algorithm specifications, direct numerical execution, exact training budgets, physics diagnostics, lineage-aware control, and effective-configuration deduplication.

\paragraph{LLM agents for scientific machine learning.}
PINNsAgent uses LLM agents and physics-guided knowledge replay to automate PDE surrogation \citep{wuwu2025pinnsagent}. Our focus is complementary: we study a population-based architecture that guides LLM design across trained PINN generations. We preserve within-run memory but do not share it across independent experimental seeds, enabling each trajectory to depend only on its own measured evolutionary feedback.

\section{Method}

\subsection{Search object and PINN objective}

Let $\mathcal{T}$ denote a PDE task containing a domain, governing laws, and physical constraints. A PINN $u_{\theta}$ is trained with
\begin{equation}
\mathcal{L}(\theta; a)=
\lambda_{r}\,\mathbb{E}_{x\sim S_a}[\|\mathcal{R}[u_{\theta}](x)\|_2^2]
+\sum_{c\in\mathcal{C}}\lambda_c\,\mathbb{E}_{x\sim C_c}[\|\mathcal{C}_c[u_{\theta}](x)\|_2^2],
\end{equation}
where the algorithm specification $a$ determines the architecture, activation, initialization, sampler $S_a$, optimizer, schedule, loss weights, and constraint transform. We optimize solution MSE on a fixed evaluation grid,
\begin{equation}
J(a)=\frac{1}{N}\sum_{i=1}^{N}\left(u_{\theta_a}(x_i)-u^{\star}(x_i)\right)^2,
\end{equation}
with relative $L_2$ error as a secondary metric and PDE/constraint residuals retained as diagnostics.

The \texttt{AlgorithmSpec} is open-world at the schema level. It contains structured fields for network, loss, sampling, optimizer, physical parameters, output transforms, and training schedule. The runtime advertises currently available components, but those components are examples rather than a discretized hyperparameter grid. Continuous choices remain LLM decisions.

\begin{figure*}[t]
    \centering
    \includegraphics[width=0.98\textwidth]{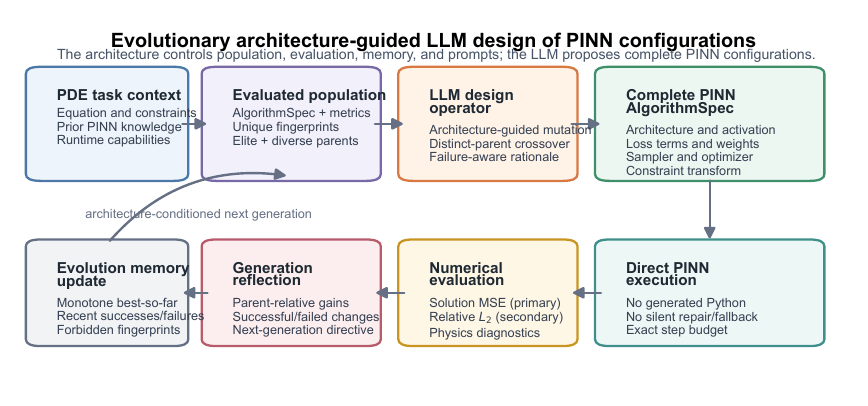}
    \caption{\textbf{Our evolutionary architecture for LLM-based PINN design.} The architecture controls the evaluated population, parent selection, elite survival, numerical evaluation, cross-generation memory, and the next prompt. The LLM acts as a conditional mutation/crossover operator over complete \texttt{AlgorithmSpec} objects. Effective fingerprints prevent metadata-only renaming from masquerading as a new algorithm.}
    \label{fig:method}
\end{figure*}

\subsection{Evolutionary architecture and cross-generation control}

At generation $g$, the population $\mathcal{P}_g$ contains evaluated specifications, metrics, and lineage. Generation zero creates unconstrained proposals. Each later generation follows the architecture's evolutionary design loop: select measured parents, construct a state-conditioned LLM prompt, execute the resulting PINNs, preserve elite and diverse survivors, and reflect on parent-relative changes. Thus, every later prompt is assembled by the architecture from earlier numerical evaluations.

For targeted mutation, the architecture supplies the LLM with one evaluated parent and requests changes to one to three fields while preserving measured strengths and avoiding recorded failures. For distinct-parent crossover, the architecture selects two parents with different effective fingerprints. One parent comes from the elite set, while the second favors configuration distance subject to competitive objective value. The generated prompt includes the PDE description, runtime capabilities, cross-generation memory, current population, parent metrics, forbidden fingerprints, and current best MSE. These inputs make each LLM proposal conditional on the search history rather than only on the PDE description.

Each response also contains explicit evolution metadata: parent identifiers, changed fields, preserved strengths, avoided failure modes, predicted improvement, and a novelty claim. These explanations do not determine fitness; they make the LLM's hypothesis auditable against the measured result.

\subsection{Effective fingerprints and candidate selection}

Candidate identifiers, names, generation indices, rationales, and lineage do not alter execution. We therefore canonicalize only effective fields and compute
\begin{equation}
h(a)=\operatorname{SHA256}\!\left(\operatorname{canonical}(a\setminus\mathcal{M})\right)_{1:16},
\end{equation}
where $\mathcal{M}$ is the set of metadata fields. A proposal is rejected if its fingerprint has already been evaluated or appears earlier in the same generation.

Novel proposals are scored before expensive training using
\begin{equation}
q(a)=0.40\hat{\Delta}(a)+0.30\nu(a)+0.15\tau(a)+0.10e(a)+0.05r(a),
\end{equation}
where $\hat{\Delta}$ is the LLM's bounded predicted improvement, $\nu$ is distance-based novelty, $\tau$ rewards targeted changes, $e$ records whether parent strengths and failures are explicitly used, and $r$ indicates a non-empty rationale. Selection cycles across proposal, mutation, and crossover buckets to avoid domination by one operator. The pool includes reserve candidates so failed executions do not consume a formal training slot.

\subsection{Population update and generation reflection}

After training, configurations are ranked by primary and secondary metrics. The best $k$ unique configurations are retained as elites. Remaining slots balance rank and minimum distance to already selected configurations, preserving alternative lineages.

For a child $a_i$ with parent set $\Pi_i$, reflection computes
\begin{equation}
\Delta_i=J(a_i)-\min_{p\in\Pi_i}J(p).
\end{equation}
A negative $\Delta_i$ records a successful hypothesis; otherwise the change becomes a failure pattern. The memory stores the previous best, generation best, monotone best-so-far, successful and failed changes, the best configuration summary, and a directive for the next generation. Memory is local to a run and is not shared across seeds.

\subsection{Direct and auditable execution}

The LLM never emits Python. Its structured specification is passed to a fixed PyTorch PINN runtime. Registered components are instantiated directly; unregistered choices fail explicitly. There is no silent repair, fallback replacement, or substitution with a hand-designed candidate. All successful models obey the same exact optimizer-evaluation budget. This contract separates the creative search operator from the trusted numerical executor.

\section{Experiments}

\subsection{Wave-C benchmark}

We study the one-dimensional constant-coefficient wave equation
\begin{equation}
u_{tt}-4u_{xx}=0,\qquad (x,t)\in[0,1]^2,
\end{equation}
with homogeneous Dirichlet boundaries and
\begin{align}
u(x,0)&=\sin(\pi x)+0.5\sin(4\pi x),\\
u_t(x,0)&=0.
\end{align}
The analytic solution is
\begin{equation}
u^{\star}(x,t)=\sin(\pi x)\cos(2\pi t)+0.5\sin(4\pi x)\cos(8\pi t).
\end{equation}
The mixed spatial frequencies make activation, width, constraint handling, and collocation density consequential while retaining exact evaluation.

\subsection{Search protocol}

We run two independent searches with seeds 0 and 1. Each run uses ten generations, population size ten, ten requested mutations and ten requested crossovers per generation, and three successful PINNs per generation. Each successful PINN receives exactly 10,000 optimizer evaluations. The primary metric is evaluation-grid MSE and the secondary metric is relative $L_2$ error. We use 64 boundary points, 64 initial points, a $32\times32$ evaluation grid, DeepSeek-chat as the LLM, and a single NVIDIA RTX 4060 Laptop GPU with PyTorch 2.5.1 and CUDA 12.4.

Across both runs, 60 successful PINNs consumed 600,000 optimizer evaluations. The search made 400 LLM calls and approximately 13.7 million language-model tokens. Aggregate PINN training time was 8.4 GPU-hours. All artifacts include specifications, parent identifiers, prompts, metrics, exact step counts, and MSE traces.

\subsection{Evaluation questions}

We ask: (1) Does the best-so-far MSE improve after generation zero? (2) Do mutation and crossover produce measured parent improvements? (3) Can the lineage reveal which parameter changes drive progress? (4) Does the MSE objective remain aligned with physics residual quality? Because only two seeds and one equation are available, we report trajectories and effect sizes without confidence intervals or significance tests.

\section{Results}

\subsection{The final generation outperforms the initial LLM proposals}

\begin{table}[t]
\caption{Independent Wave-C searches. Reduction is relative to the best generation-zero configuration.}
\label{tab:main-results}
\centering
\small
\begin{tabular}{lrrrrr}
\toprule
Run & Initial MSE & Final MSE & Reduction & Rel. $L_2$ & Best gen.\\
\midrule
Seed 0 & $6.569\!\times\!10^{-2}$ & $6.374\!\times\!10^{-2}$ & 2.97\% & 0.4518 & 9\\
Seed 1 & $6.521\!\times\!10^{-2}$ & $3.015\!\times\!10^{-3}$ & 95.38\% & 0.0983 & 9\\
\midrule
Mean & -- & $3.338\!\times\!10^{-2}$ & -- & 0.2751 & --\\
\bottomrule
\end{tabular}
\end{table}

\begin{figure*}[t]
    \centering
    \includegraphics[width=0.98\textwidth]{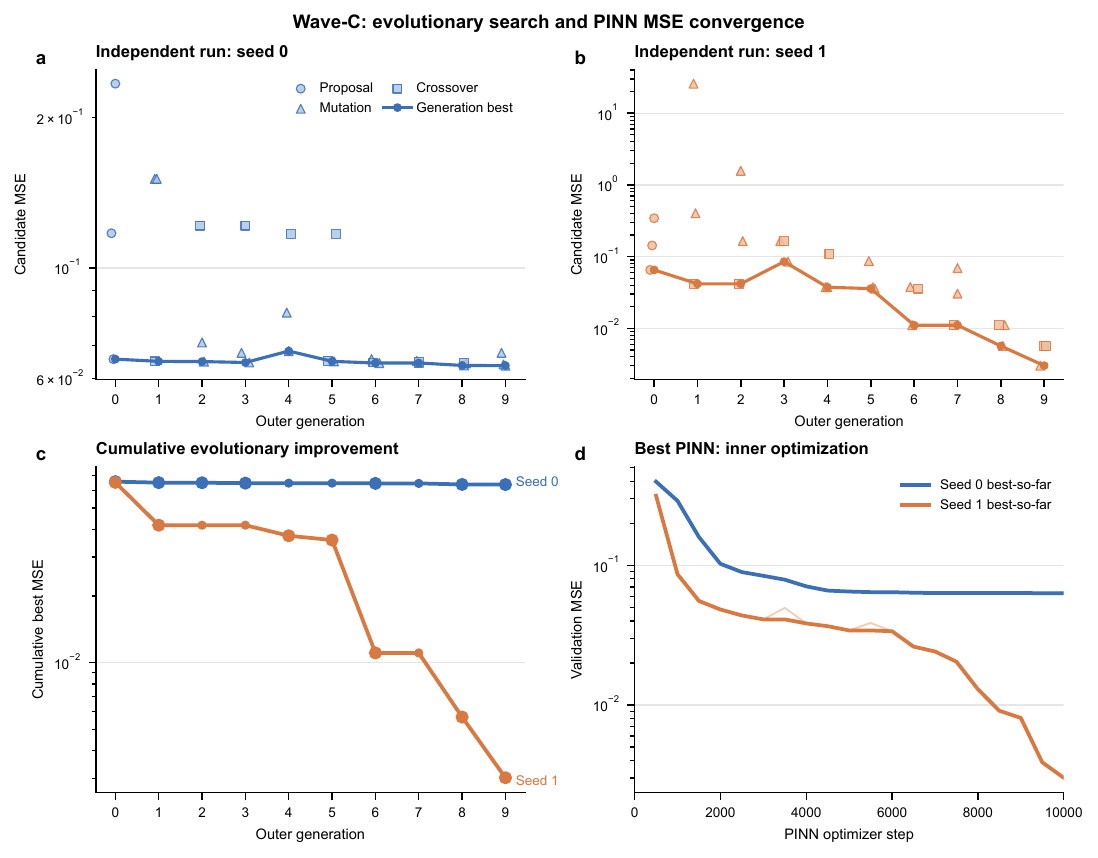}
    \caption{\textbf{Search and optimization trajectories.} (a,b) Candidate MSE by generation for the two independent runs; the solid line is the generation best. (c) Monotone cumulative-best MSE. (d) Best-so-far validation MSE within the 10,000 optimizer steps of each run's final selected PINN. The candidate scatter includes two post-audit contract controls described in Appendix~\ref{app:audit}; neither changed a historical elite.}
    \label{fig:search-results}
\end{figure*}

Table~\ref{tab:main-results} and Figure~\ref{fig:search-results} show that neither independent run was solved by its initial population. Seed 0 improved modestly but repeatedly: its best-so-far MSE decreased in generations 1, 2, 3, 6, 8, and 9. Seed 1 displayed a stronger trajectory, reducing MSE from $0.0652$ to $0.00302$. In both cases the minimum appeared in generation 9. The key observation is temporal: the strongest configurations were produced only after our architecture had accumulated measured population history, not in the LLM's initial proposals.

\subsection{Cross-generation memory combines successful changes}

In seed 1, a generation-1 crossover combined the four-layer, 64-unit Swish MLP from the best initial proposal with equal loss weights and an initial-boundary output transform. MSE fell from $0.06521$ to $0.04175$, a 36.0\% reduction. Later crossovers did not set new records in this run, so the evidence supports one useful early recombination but not a general advantage for crossover.

The clearest architecture-level adaptation occurred across generations 4 and 5. Two generation-4 mutations independently provided positive evidence for residual connections and depth five. The architecture placed both measured successes in the generation-5 context. The LLM then took the depth-five elite as its parent and applied the recorded residual-network change, reducing MSE to $0.03577$. This is a concrete instance of cross-generation memory combining knowledge from separate branches. Subsequent targeted mutations widened the resulting residual MLP from 64 to 128 units, then to 256 units, and finally increased collocation from 256 to 512 points. Their parent-relative MSE reductions were 69.1\%, 48.7\%, and 46.8\%, respectively. The final configuration was therefore accumulated through architecture-guided numerical updates rather than emitted as a single guess.

\begin{table}[t]
\caption{Measured seed-1 evolutionary improvement sequence. The two generation-4 rows are separate successful branches; generation 5 combines their validated changes through cross-generation memory. All rows use Swish, Adam with learning rate $10^{-3}$, no scheduler, equal loss weights, and the initial-boundary transform unless noted.}
\label{tab:lineage}
\centering
\small
\begin{tabular}{rllrr}
\toprule
Gen. & Evolutionary change & Network & Colloc. & MSE\\
\midrule
0 & Initial proposal & MLP $4\times64$ & 256 & 0.065209\\
1 & Crossover: weights + transform & MLP $4\times64$ & 256 & 0.041751\\
4 & Branch A: residual blocks & ResMLP $4\times64$ & 256 & 0.037490\\
4 & Branch B: depth $4\rightarrow5$ & MLP $5\times64$ & 256 & 0.037408\\
5 & Reflection combines A + B & ResMLP $5\times64$ & 256 & 0.035773\\
6 & Mutation: width $64\rightarrow128$ & ResMLP $5\times128$ & 256 & 0.011051\\
8 & Mutation: width $128\rightarrow256$ & ResMLP $5\times256$ & 256 & 0.005672\\
9 & Mutation: collocation $256\rightarrow512$ & ResMLP $5\times256$ & 512 & 0.003015\\
\bottomrule
\end{tabular}
\end{table}

The cross-generation memory also retained negative information. Step learning-rate schedules, residual-adaptive sampling, depth six, and several wider or deeper non-residual networks increased MSE. Elite preservation prevented these candidates from replacing the best configuration, while reflection exposed their changed fields and parent-relative failures to later prompts.

\subsection{Inner optimization preserves the outer-loop gains}

The best-so-far curves in Figure~\ref{fig:search-results}d decrease monotonically within each selected PINN. Seed 0 largely plateaus after approximately 6,000 steps, whereas seed 1 continues to improve through the final evaluation interval. Thus the outer-loop gains are not artifacts of reporting only a transient early checkpoint; each model receives the same full training budget.

\subsection{MSE-only evolution exposes objective mismatch}

The MSE-optimal seed-1 model has relative $L_2$ error 0.0983, essentially zero boundary and initial-condition errors, and PDE residual error 8.12. By contrast, the best seed-0 model has MSE 0.0637 and PDE residual 0.109. The seed-1 lineage uses an output transform that exactly enforces the boundary and initial conditions, while the MSE objective rewards a wide residual network that approximates the analytic solution on the evaluation grid. The high residual shows that solution accuracy and differential consistency are not interchangeable objectives.

This is not evidence that the evolutionary loop failed: it optimized the declared primary metric. It is evidence that a scientific search system must make the objective contract physics-aware. A practical extension is constrained or Pareto evolution, e.g., minimizing MSE subject to a residual threshold, or maintaining a front over MSE, relative $L_2$, and governing-law residual.

\section{Limitations and Scope}

The principal limitation is experimental breadth. Wave-C is one forward, analytic, one-dimensional PDE; two seeds are insufficient for variance estimates. We do not compare against Bayesian optimization, random search, population-based training, OPRO, or PINNsAgent under matched compute. The operator analysis is lineage-resolved but observational, not a factorial ablation. It therefore shows that the architecture used measured evolutionary feedback, but does not isolate the causal contribution of each component. DeepSeek-chat is the only LLM evaluated, so model dependence is unknown.

The experiment also revealed an implementation issue during strict final audit: two generations in seed 1 produced only two unique post-deduplication candidates. After fixing the refill logic, we trained two conservative control candidates to complete the 30-model accounting. Both were worse than their historical generation elites and therefore did not change any best-so-far value or later lineage; details are in Appendix~\ref{app:audit}. A clean rerun with the patched refill mechanism is nevertheless required before a definitive benchmark claim.

For a full ICLR evaluation, the method should be tested on multiple PINNacle equations, including elliptic, parabolic, hyperbolic, and nonlinear conservation-law problems; at least five seeds; matched evaluation budgets; LLM and non-LLM search baselines; and ablations removing reflection memory, fingerprint deduplication, distinct-parent crossover, elite protection, and diversity preservation.

\section{Conclusion}

We presented a custom evolutionary architecture that guides an LLM to design complete PINN configurations through population search rather than isolated recommendation. The architecture supplies the optimization logic: parent selection, lineage, targeted mutation, distinct-parent crossover, elite preservation, deduplication, numerical evaluation, and cross-generation memory. In the Wave-C study, both runs improved beyond their initial populations. One run used measured feedback to combine successful structural changes and refine width and collocation density, reducing MSE by 95.38\%. The same experiment exposed a physics-objective mismatch, defining a concrete next step: multi-objective, physics-audited evolution across diverse PDE benchmarks.

\bibliography{references}
\bibliographystyle{iclr2026_conference}

\appendix

\section{AlgorithmSpec summary}
\label{app:algorithmspec}

The structured search object contains: (i) network type, depth, width, activation, initialization, and component parameters; (ii) named loss terms with categories, physical targets, weights, and weighting strategy; (iii) sampling method and collocation count; (iv) optimizer, learning rate, scheduler, and component-specific parameters; (v) physical-constraint modes and optional output transform; (vi) trainable physical parameters; and (vii) training schedule. Lineage and design rationale are stored as metadata but removed from the effective fingerprint.

\section{Per-generation best MSE}

\begin{table}[h]
\caption{Generation-best and cumulative-best MSE.}
\centering
\small
\begin{tabular}{rrrrr}
\toprule
& \multicolumn{2}{c}{Seed 0} & \multicolumn{2}{c}{Seed 1}\\
Gen. & Generation & Cumulative & Generation & Cumulative\\
\midrule
0 & 0.065693 & 0.065693 & 0.065209 & 0.065209\\
1 & 0.065014 & 0.065014 & 0.041751 & 0.041751\\
2 & 0.064954 & 0.064954 & 0.041751 & 0.041751\\
3 & 0.064676 & 0.064676 & 0.085107 & 0.041751\\
4 & 0.068181 & 0.064676 & 0.037408 & 0.037408\\
5 & 0.065026 & 0.064676 & 0.035773 & 0.035773\\
6 & 0.064509 & 0.064509 & 0.011051 & 0.011051\\
7 & 0.064542 & 0.064509 & 0.011051 & 0.011051\\
8 & 0.063808 & 0.063808 & 0.005672 & 0.005672\\
9 & 0.063739 & 0.063739 & 0.003015 & 0.003015\\
\bottomrule
\end{tabular}
\end{table}

\section{Strict-budget and recovery audit}
\label{app:audit}

The final artifact audit contains 30 successful PINNs per seed. Every successful model records exactly 10,000 optimizer evaluations and an exact-budget flag. Seed 0 required one additional proposal after an execution failure. In seed 1, the original online search trained 28 successful models because generations 5 and 7 each exhausted their unique proposal pools after two candidates. The patched search now generates supplemental unique proposals and maintains failure reserves. For accounting completeness, two post-hoc control configurations (three-layer tanh MLPs of width 32 and 48) were trained for 10,000 steps and assigned to the missing generations. Their MSE values were 0.086806 and 0.069211, respectively; neither changed a population ranking, best-so-far trajectory, or reported lineage. Operator-effect statements in the main text exclude these controls.

\section{Reproducibility artifacts}

The project records the complete search configuration, all LLM traces with secrets redacted, generation memories, candidate fingerprints, parent identifiers, structured specifications, numerical metrics, loss histories, MSE traces, environment metadata, and exact optimizer counts. The formal experiment used Python 3.11.5, PyTorch 2.5.1+cu124, CUDA 12.4, Windows 10, and one NVIDIA RTX 4060 Laptop GPU. Independent seeds do not share evolutionary memory.

\end{document}